\documentclass[11pt]{article}
\usepackage{booktabs}
\usepackage{multirow}
\usepackage{xcolor}
\usepackage{pifont}
\usepackage{amsmath}
\definecolor{risegreen}{RGB}{0,150,0}
\definecolor{fallgray}{RGB}{120,120,120}
\definecolor{neutralgray}{RGB}{140,140,140}
\usepackage[table]{xcolor} 
\usepackage{amssymb} 
\usepackage{wrapfig}
\usepackage{placeins}
\usepackage{subcaption}
\usepackage[most]{tcolorbox}
\usepackage{enumitem}
\usepackage{algorithm}
\usepackage{algpseudocode}

\definecolor{risegreen}{RGB}{0,150,0}
\definecolor{deltagray}{gray}{0.45}

\providecommand{\rise}[1]{}
\providecommand{\fall}[1]{}
\providecommand{\neutral}[1]{}

\renewcommand{\rise}[1]{\textcolor{risegreen}{\,(\,+#1\,)}}
\renewcommand{\fall}[1]{\textcolor{deltagray}{\,(\,-#1\,)}}
\renewcommand{\neutral}[1]{\textcolor{deltagray}{\,(\ensuremath{\pm}#1\,)}}

\usepackage[preprint]{acl}

\usepackage{times}
\usepackage{latexsym}

\usepackage[T1]{fontenc}

\usepackage[utf8]{inputenc}

\usepackage{microtype}

\usepackage{inconsolata}

\usepackage{graphicx}

\title{Reducing Peak Memory Usage for \\ Modern
Multimodal Large Language Model Pipelines}

\author{
Junwan Kim\textsuperscript{*}
\and
Hyunkyung Bae\textsuperscript{*}\\
New York University \\
\texttt{\{junwan.kim, hyunkyung.bae\}@nyu.edu}
}

\definecolor{lightgray}{rgb}{0.7,0.7,0.7}

\begin{document}
\maketitle
\begingroup
\renewcommand{\thefootnote}{*}
\footnotetext{Equal Contribution}
\endgroup
\begin{abstract}
Multimodal large language models (MLLMs) achieve strong visual--textual reasoning by scaling to high-resolution images and long video sequences, but this scalability introduces substantial inference-time memory overhead due to the growth of the key--value (KV) cache. Existing KV-cache compression methods primarily operate after the full multimodal context has been processed, and therefore do not address the peak memory consumption incurred during the prefill stage. We observe that visual tokens in MLLMs exhibit strong structural regularities and representational redundancy that can be exploited earlier in the inference pipeline. Based on this observation, we propose a sequential, structure-aware KV-cache compression framework that operates during prefill and enforces a fixed memory budget throughout input processing. Unlike conventional post-prefill compression, which first constructs the full KV cache and compresses it afterward, our method compresses incrementally during prefix encoding. Experimental results show that our approach substantially reduces peak memory usage with minimal degradation in generative performance, enabling more practical and memory-efficient multimodal inference for large-scale visual inputs.
\end{abstract}

\section{Introduction}
Multimodal large language models (MLLMs) have emerged as a powerful paradigm for jointly reasoning over visual and textual inputs, enabling applications such as visual question answering~\cite{antol2015vqa}, image-based reasoning~\cite{shen2025vlmr1}, and video understanding~\cite{zhang2024llavavideo}. To support these capabilities, modern MLLMs process increasingly complex visual signals, ranging from single images to high-resolution tiled patches and long video sequences. In a typical architecture~\citep{liu2023llava}, a pretrained vision encoder extracts visual features, an adaptor projects them into the language embedding space, and a transformer backbone jointly attends over vision tokens and textual inputs. While this unified attention enables flexible multimodal integration, it introduces substantial computational and memory challenges as the number of input tokens grows.

A key bottleneck arises from the self-attention operation~\cite{vaswani2017attention}, whose complexity scales quadratically with sequence length. Autoregressive transformers alleviate this cost through key--value (KV) caching~\cite{pope2023kvcache}, which stores intermediate attention representations and reduces per-token decoding complexity from $\mathcal{O}(N^2)$ to $\mathcal{O}(N)$. However, KV caching introduces a severe memory burden: the cache grows linearly with the number of tokens and must be retained across all layers and attention heads.

This challenge is particularly acute in multimodal settings. Recent advances in MLLMs have been driven by aggressively increasing the number of vision tokens, including tiled representations for high-resolution images~\cite{bai2023qwen, chen2024internvl, tong2024cambrian-1}, dense frame sampling for videos~\cite{xu2024pllava, zhang2024llavavideo, yang2025cambrian-s}, and multi-view visual inputs~\cite{cheng2025sr3d, huang2025mllm}. These design choices substantially inflate the token count before decoding begins, causing the KV cache constructed during input processing to dominate memory usage. As a result, the prefill stage—where the full multimodal prefix is encoded—becomes the point of peak memory consumption during inference.

Prior work has sought to reduce inference-time memory usage primarily through KV-cache compression~\cite{li2024snapkv, kim2025kvzip, wan2025meda}. These methods exploit redundancy by evicting, merging, or approximating cached key--value pairs and are effective for long-context decoding. However, they typically apply compression only after the entire multimodal context has been processed, leaving the peak memory spike during prefill unaddressed. Conceptually, existing methods follow a \emph{process first, compress later} pipeline, whereas our approach follows a \emph{compress as you prefill} pipeline to keep memory bounded throughout input processing. Token pruning methods~\cite{yang2025visionzip, zhang2025beyond} reduce memory by discarding input tokens, but operate at the input level and ignore the heterogeneous roles that different layers and attention heads assign to tokens~\cite{yoon2025visual, zhang2025cross, kaduri2025s}, increasing the risk of removing structurally important information.

In this work, we argue that the prefill stage itself offers untapped opportunities for memory-efficient multimodal inference. Visual inputs exhibit strong structural regularities: images consist of spatially coherent regions, and videos contain substantial temporal redundancy across frames. These structures form coarse-to-fine representations of the same underlying content, and not all visual tokens contribute equally to downstream reasoning.

Motivated by this observation, we propose a prefill-aware, structure-aware (i.e., aware of the spatial and temporal structure and redundancy of visual tokens) KV-cache compression framework that operates sequentially under a fixed memory budget. For single-turn settings, we introduce a query-aware strategy that leverages the textual prompt during prefill to estimate token importance and retain visually salient regions. For potential multi-turn interactions, where query signals may be unavailable, we explore a query-agnostic variant that relies solely on the structural and representational properties of visual tokens. Together, these approaches substantially reduce peak memory usage during inference while preserving downstream performance, enabling scalable and memory-efficient multimodal inference across diverse interaction patterns.
\section{Preliminaries}

\subsection{KV Cache in Transformer Inference}
Transformers~\cite{vaswani2017attention}, as used in large language models~\cite{brown2020gpt3}, generate tokens autoregressively. At each step, self-attention computes interactions between the current query and all previously generated tokens. For a sequence of length $t$, attention is defined as:
\begin{equation}
\text{Attention}(Q, K, V) = \text{softmax}\!\left(\frac{QK^{\top}}{\sqrt{d_k}}\right)V ,
\end{equation}
where $Q$, $K$, and $V$ denote query, key, and value matrices, and $d_k$ is the key dimension. During generation, only the query for the current token is newly computed, while keys and values from all preceding tokens are reused. To avoid recomputation, these keys and values are stored in GPU memory as a key--value (KV) cache.

\subsection{The Necessity of KV Cache Management}
KV caching reduces per-token decoding complexity from $\mathcal{O}(N^2)$ to $\mathcal{O}(N)$, but introduces a memory overhead that scales linearly with sequence length and model size. The total KV-cache memory footprint can be approximated as:
\begin{equation}
\begin{aligned}
\text{Memory}_{\text{KV}}
\approx\; &2 \times \text{layers} \times \text{heads} \times \text{dim}_{\text{head}} \\
&\times \text{precision} \times \text{sequence length} ,
\end{aligned}
\end{equation}
where the factor of $2$ accounts for both keys and values. As models scale to ultra-long contexts (e.g., $100$K$+$ tokens), the KV cache alone can exceed available GPU memory, making effective KV-cache management essential for inference under fixed memory budgets.

\subsection{The Vision Token Explosion Problem}
Modern multimodal large language models (MLLMs) support high-resolution images and long video sequences through dense visual tokenization, resulting in substantially longer input sequences than text-only models. High-resolution images are decomposed into spatial grids of patches, each represented as a vision token. For an image of resolution $H \times W$, the number of vision tokens is:
\begin{equation}
N_{\text{vis}} = \frac{H \times W}{P^2} ,
\end{equation}
where $P$ is the patch size. For example, an $4$K image ($3840 \times 2160$) yields over $42{,}000$ vision tokens with $P=14$, demonstrating how visual inputs can dominate the token budget before decoding begins and drive peak memory usage during prefill.

\section{Methodology}
\begin{algorithm}[t]
\caption{Block-wise Prefill with KV Eviction}
\label{alg:block_prefill}
\begin{algorithmic}[1]
\State \textbf{Input:} Input sequence $S$, Block size $b$, Memory budget $M$
\State \textbf{Output:} Compressed KV Cache $\mathcal{C}$
\State Partition $S$ into blocks $\{B_1, B_2, \dots, B_N\}$ of size $b$
\State $\mathcal{C} \gets \emptyset$ \Comment{Initialize empty cache}

\For{each block $B_i \in \{B_1, \dots, B_N\}$}
    \State $(K_i, V_i) \gets \text{ComputeKV}(B_i)$ \Comment{Generate KV pairs for current block}
    \State $\mathcal{C} \gets \mathcal{C} \cup (K_i, V_i)$ \Comment{Append new pairs to cache}
    
    \If{$|\mathcal{C}| > M$}
        \State $k_{excess} \gets |\mathcal{C}| - M$
        \State $\mathcal{C} \gets \text{Evict}(\mathcal{C}, k_{excess})$ \Comment{Reduce cache to budget $M$}
    \EndIf
\EndFor
\State \Return $\mathcal{C}$
\end{algorithmic}
\label{alg:block_prefill}
\end{algorithm}
We propose a prefill-aware inference framework that reduces peak memory usage in multimodal large language models (MLLMs) by enforcing a fixed KV-cache budget throughout input processing, rather than compressing the cache only after the full multimodal context has been encoded.

\subsection{Block-wise Processing for MLLMs}
Conventional KV-cache eviction strategies construct the full KV cache before pruning, leading to high peak memory usage and frequent out-of-memory failures during prefill—particularly in MLLMs, where high-resolution images and long videos introduce thousands of vision tokens.

To address this issue, we adopt block-wise prefill~\cite{kim2024infinipot, kim2025epicache}, partitioning the input sequence into contiguous blocks that are processed sequentially, as summarized in Alg.~\ref{alg:block_prefill}. After each block is encoded, its KV pairs are appended to the cache and pruned to satisfy a fixed budget $M$. This explicitly bounds the KV-cache size throughout prefill, preventing peak memory growth.

Block-wise prefill is well suited to multimodal inputs. Unlike text, visual inputs exhibit strong structural organization: images consist of spatially coherent tiles, and videos of temporally contiguous frame groups. We align block boundaries with these visual structures, enabling eviction decisions to be made at semantically meaningful granularity and improving robustness to compression.

\subsection{Eviction Strategies}
Within the block-wise framework, we consider two complementary eviction strategies that differ in their reliance on query information. Both operate online during prefill and are applied immediately after each block.

\paragraph{Query-Aware Eviction.}
For single-turn settings, we adopt a query-aware eviction strategy based on SnapKV~\cite{li2024snapkv}. Proxy query tokens are extracted from the textual prompt and used to compute cross-attention over cached keys. Given query features $q_{\text{obs}}$ and cached key $k_j$, the importance score is:
\vspace{-1pt}
\begin{equation}
\alpha_j = \text{Softmax}\!\left(\frac{q_{\text{obs}} \cdot k_j^{\top}}{\sqrt{d_k}}\right) .
\end{equation}
\vspace{-1pt}
Tokens with lower importance scores are evicted until the cache satisfies the budget $M$. Applied sequentially during prefill, this strategy prioritizes visually salient regions relevant to the task while discarding redundant tokens early.

\paragraph{Query-Agnostic Eviction.}
For potential multi-turn scenarios where query signals may be unavailable, we employ a query-agnostic strategy based on KeyDiff~\cite{keydiff}. The method preserves representational diversity by retaining keys that deviate most from the average representation. Specifically, we define an anchor vector $\mu$ as the mean of cached keys and prioritize retention of keys with lower similarity to $\mu$. This avoids $\mathcal{O}(N^2)$ pairwise comparisons while preserving outliers and rare visual features without relying on query information.


\section{Experiments}
\label{sec:exp}

\subsection{Experimental Setup}
\label{subsec:exp_setup}
\paragraph{Benchmarks.} We evaluate on benchmarks that are sensitive to the scale and structure of visual tokens. For images, we use ImageNeedleInHaystack from MileBench~\cite{song2024milebench} and V$^*$~\cite{wu2024vstar}, which require dense visual localization. For videos, we adopt MLVU~\cite{zhou2024mlvu} and the long-video setting of Video-MME~\cite{fu2025video}. All experiments are conducted on NVIDIA A100 GPUs using standard evaluation protocols. We report task accuracy, average accuracy, and the difference $\Delta$ relative to the full-cache baseline.
\vspace{-3pt}

\paragraph{Models and settings.}
\begin{table}[t]
\centering
\resizebox{0.48\textwidth}{!}{
\setlength{\tabcolsep}{2pt}
\begin{tabular}{c|cccccc}
\toprule[1.25pt]
\toprule 
Method (KV Budget) & ImageNeedle & V$^*$ & MLVU & Video-MME (L) & Average & $\Delta$ \\
\midrule
\multicolumn{7}{l}{\textbf{\textit{\small{InternVL3.5-8B}}}} \\
\midrule
Full Cache   &  80.31 & 84.35 & 51.28 & 53.89 & 67.46 & - \\
\midrule
SnapKV (4096)  & 80.94 & 82.72 & 50.00 & 52.33 & 66.50 & 0.96  \\
SnapKV (2048)  & 80.31 & \textbf{83.76} & 49.61 & 52.33 & 66.50 & 0.96  \\
SnapKV (1024)  & 80.00 & 82.61 & \textbf{51.00 }& \textbf{53.11} & \textbf{66.68} & \textbf{0.78}  \\
KeyDiff (4096)         & \textbf{83.13} & 74.87 & 49.60 & 51.33 & 64.03 & 3.43 \\
KeyDiff (2048)         &  79.69 & 75.39 & 50.40 & 52.00 & 65.23 & 2.23 \\
KeyDiff (1024)         & 74.06 & 74.35 & 50.40 & 52.22 & 62.76 & 4.70 \\
\midrule
\multicolumn{7}{l}{\textbf{\textit{\small{Qwen2.5-VL-7B}}}} \\
\midrule
Full Cache  & 83.70 & 79.58  & 48.80 & 50.00 & 65.52 & - \\
\midrule
SnapKV (4096) & \textbf{85.00} & 78.53 & 44.82 & 48.77 & 64.28 & 1.24  \\
SnapKV (2048) & 72.19 & 78.53 & 45.42 & 49.11 & 61.31 & 4.21  \\
SnapKV (1024) & 45.31 & 76.96 & 46.02 & \textbf{49.56} & 54.46 & 11.06  \\
KeyDiff (4096)         & 81.56 & \textbf{79.58} & \textbf{47.41} & 49.33 & \textbf{64.47} & \textbf{1.05}  \\
KeyDiff (2048)         & 78.44 & 69.63 & 46.41 & 48.00 & 60.62 & 4.9  \\
KeyDiff (1024)         & 66.25 & 67.02 & 43.43 & 46.55 & 55.82 & 9.7  \\
\bottomrule[1.25pt] 
\end{tabular}
}
\caption{\textbf{Performance under fixed KV-cache budgets.} Best results are in bold. $\Delta$ denotes the difference from the full-cache baseline. Our method maintains stable performance across compression settings, with minimal degradation even at a budget of 1024 ($\sim90\%$ compress)}

\label{tab:main_result_trimmed}
\end{table}
\begin{figure}[ht]
    \centering
    \includegraphics[width=0.95\linewidth]{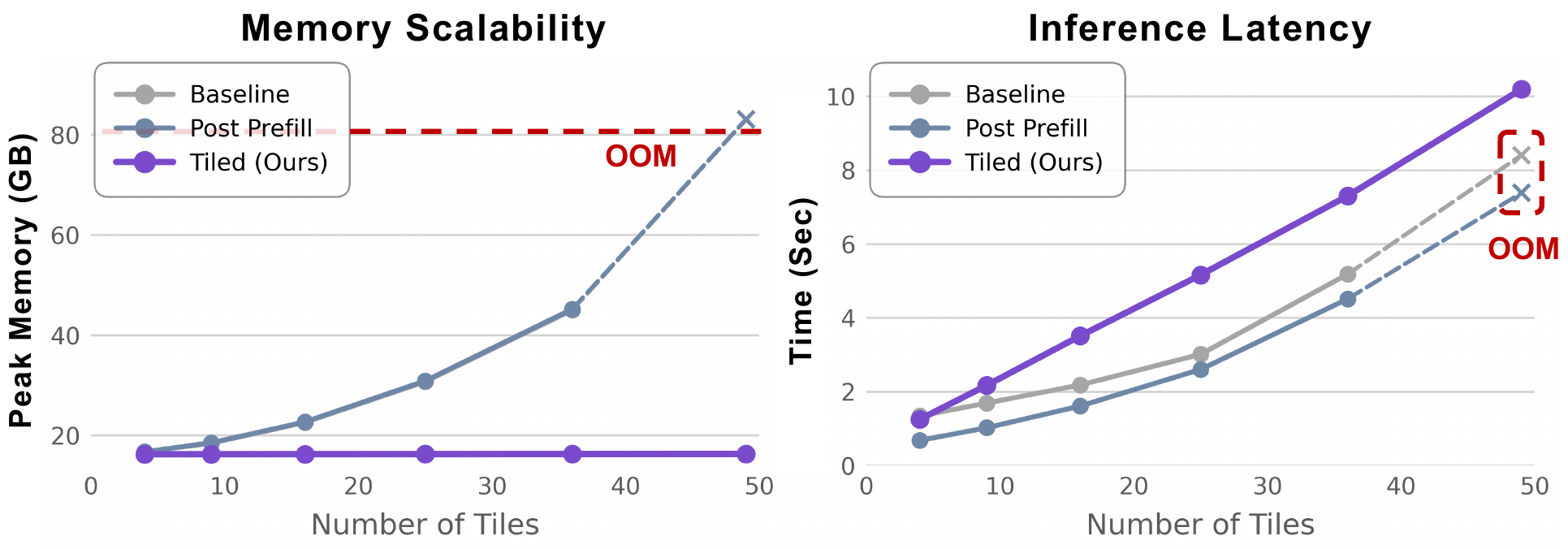}
    \caption{\textbf{Peak memory usage and inference latency} as the number of image tiles increases (InternVL-3.5). Our method maintains nearly constant peak memory during prefill under a fixed KV-cache budget, preventing out-of-memory (OOM), at the cost of increased inference latency due to sequential processing. Dashed lines indicate estimated values for an unmeasured region.}
    \label{fig:memory_usage}
    \vspace{-10pt}
\end{figure}

We mainly evaluate InternVL3.5-8B~\cite{wang2025internvl35} and Qwen2.5-VL-7B~\cite{bai2025qwen25}, as they are among the most capable open-source multimodal models currently available and both support video inputs as well as tiling for high-resolution images. InternVL3.5-8B is tested with up to 36 image tiles (9{,}216 vision tokens) and 32 video frames (8{,}192 tokens), while Qwen2.5-VL-7B uses up to 8{,}192 vision tokens for both modalities. Unless stated otherwise, the block size is 256.

\subsection{Main Results}

As shown in Tab.~\ref{tab:main_result_trimmed}, our prefill-stage compression preserves performance under aggressive KV-cache budgets, achieving up to $\sim$90\% compression. Fig.~\ref{fig:memory_usage} shows that peak KV-cache memory remains nearly constant as image tiles increase, whereas baseline methods grow linearly and encounter out-of-memory failures beyond 36 tiles. These results demonstrate that our method controls peak memory usage during prefill without sacrificing accuracy.

\paragraph{Generalization Across Model Sizes.} 
We further evaluate our method across different model sizes, including InternVL3.5-14B and Qwen2.5-VL-32B, to assess robustness under varying capacities. 
As shown in Tab.~\ref{tab:model_scale}, our method maintains stable performance with limited degradation under aggressive KV-cache compression, indicating that the effectiveness of our approach does not depend on model size. 

\begin{table}[t]
\centering
\resizebox{0.48\textwidth}{!}{
\setlength{\tabcolsep}{2pt}
\begin{tabular}{c|cccccc}
\toprule[1.25pt]
\toprule 
Method (KV Budget) & ImageNeedle & V$^*$ & MLVU & Video-MME (L) & Average & $\Delta$ \\
\midrule
\multicolumn{7}{l}{\textbf{\textit{\small{InternVL3.5-14B}}}} \\
\midrule
Full Cache   &  84.06 & 83.76 & 49.67 & 57.89 & 68.95 & - \\
\midrule
SnapKV (1024)  & 66.25 & \textbf{82.72} & \textbf{47.61} & \textbf{56.45} & \textbf{63.26} & \textbf{7.73}  \\
KeyDiff (1024) & \textbf{75.94} & 64.92 & 46.41 & 52.78 & 60.01 & 8.84 \\
\midrule
\multicolumn{7}{l}{\textbf{\textit{\small{Qwen2.5-VL-32B}}}} \\
\midrule
Full Cache  & 96.56 & 83.77  & 48.40 & 55.22 & 70.99 & - \\
\midrule
SnapKV (1024) & 66.25 & \textbf{82.72} & \textbf{47.61} & \textbf{56.45} & \textbf{63.26} & \textbf{7.73}  \\
KeyDiff (1024) & \textbf{67.19} & 72.25 & 45.82 & 54.56 & 59.96 & 11.03  \\
\bottomrule[1.25pt] 
\end{tabular}
}
\caption{\textbf{Model Scale.} Best results are in bold. $\Delta$ denotes the difference from the full-cache baseline. Our method maintains stable performance across compression settings, with minimal degradation even at a budget of 1024 ($\sim90\%$ compress)}
\vspace{-5pt}
\label{tab:model_scale}
\end{table}

\begin{figure}[t]
    \centering
    \includegraphics[width=0.95\linewidth]{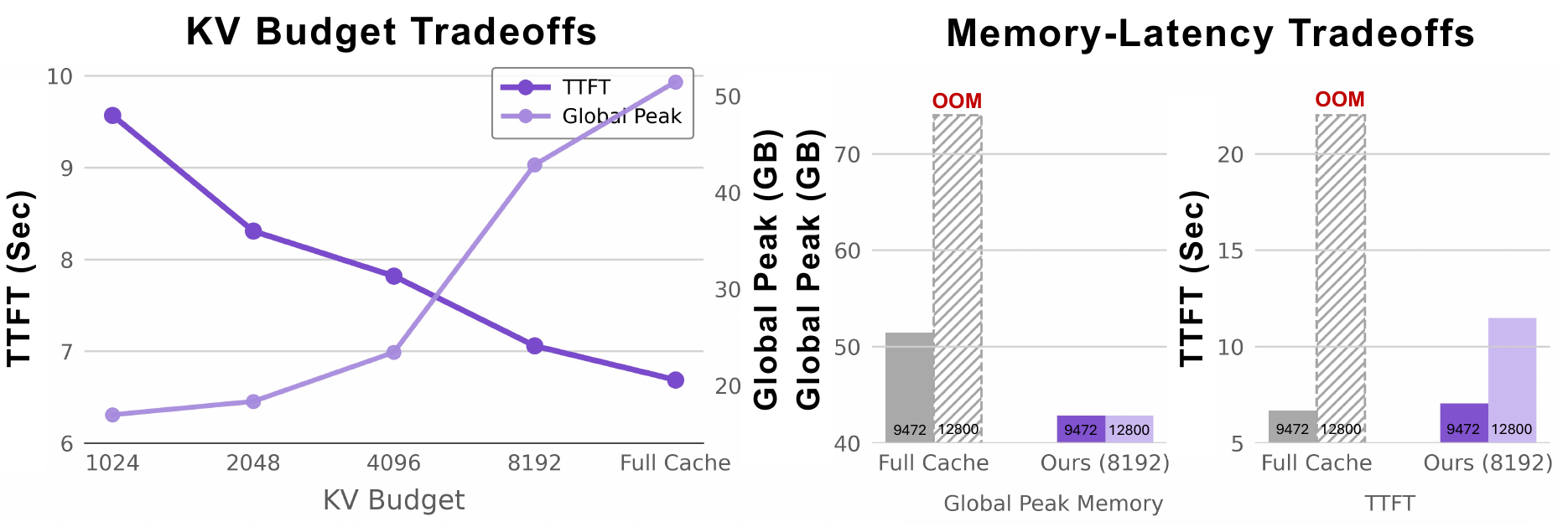}
    \caption{\textbf{Memory--latency trade-offs} under KV-cache budgeting. Left: for 9,472 vision tokens, a larger KV-cache budget reduces TTFT but increases global peak memory. Right: as the number of vision tokens increases from 9,472 to 12,800, this trade-off becomes critical---full-cache execution enters the OOM regime, whereas our method keeps memory bounded and remains executable, with a moderate TTFT increase.}
    \label{fig:peak_and_latency}
    \vspace{-10pt}
\end{figure}
\begin{table*}[t]
\centering

\begin{subtable}[t]{0.32\textwidth}
\centering
\caption{Forward under budget}
\resizebox{\textwidth}{!}{
\setlength{\tabcolsep}{12pt}
\begin{tabular}{c|c}
\toprule[1.25pt]
\toprule
Method (KV Budget) & ImageNeedle \\
\midrule
Block Forward (1024) & 80.94 \\
Bulk Forward (1024)  & 80.31 \\
\bottomrule[1.25pt]
\end{tabular}
}
\end{subtable}
\hfill
\begin{subtable}[t]{0.32\textwidth}
\centering
\caption{Static vs.\ Dynamic}
\resizebox{\textwidth}{!}{
\setlength{\tabcolsep}{12pt}
\begin{tabular}{c|c}
\toprule[1.25pt]
\toprule
Method (KV Budget) & ImageNeedle \\
\midrule
Static (1024)  & 80.31 \\
Dynamic (1024) & 74.68 \\
\bottomrule[1.25pt]
\end{tabular}
}
\end{subtable}
\hfill
\begin{subtable}[t]{0.32\textwidth}
\centering
\caption{Input res. vs.\ Compression}
\resizebox{\textwidth}{!}{
\setlength{\tabcolsep}{12pt}
\begin{tabular}{c|c}
\toprule[1.25pt]
\toprule
Method (KV Budget) & ImageNeedle \\
\midrule
Compression (1024) & 80.31 \\
Reduction (1024) & 9.38 \\
\bottomrule[1.25pt]
\end{tabular}
}
\end{subtable}
\caption{\textbf{Analysis of prefill strategies under a fixed KV-cache budget.}
(a) Forward execution strategies, (b) static vs. dynamic budgeting, and (c) input resolution reduction vs. prefill-stage compression.}
\vspace{-10pt}
\label{tab:abl}
\end{table*}

\paragraph{Memory--latency trade-off.}
Fig.~\ref{fig:peak_and_latency} illustrates the trade-off induced by our prefill-stage compression. On the left, for an input with 9,472 vision tokens, increasing the KV-cache budget reduces time to first token (TTFT) but increases global peak memory, while smaller budgets better bound memory at the cost of higher TTFT. This trade-off becomes particularly important as the number of input vision tokens grows. As shown on the right, when the input increases from 9,472 to 12,800 vision tokens, the full-cache baseline enters the out-of-memory (OOM) regime, whereas our method maintains a much lower global peak memory and remains executable, with only a moderate increase in TTFT. These results show that our method converts otherwise infeasible large-vision inputs into a controllable memory--latency trade-off.




\section{Analysis}
\label{sec:analysis}

\paragraph{Query-aware vs.\ query-agnostic eviction.}
Query-aware eviction (SnapKV) achieves the strongest performance when query signals are available, particularly at small budgets. On InternVL3.5-8B, SnapKV at budget 1024 incurs only a 0.78 average accuracy drop. The query-agnostic KeyDiff variant remains competitive, with reasonable degradation even at small budgets (e.g., 4.70 at 1024), indicating that preserving representational diversity alone retains task-relevant information and supports multi-turn settings.

\paragraph{Video tasks and non-monotonic behavior.}
On video benchmarks, reducing the cache budget does not always degrade performance monotonically. For InternVL3.5-8B, SnapKV at budget 1024 achieves slightly improved results on MLVU and Video-MME, suggesting that prefill-stage compression suppresses redundant temporal information and yields more focused representations.

\paragraph{Budgeting.}
Block-wise prefill improves memory efficiency but increases latency due to sequential execution. To reduce this overhead, we use a hybrid strategy that maximizes the amount of computation done in a single forward pass (bulk forward) and applies block-wise processing only when necessary. With a budget of 1024, processing the first $M$ tokens in one pass achieves comparable accuracy (80.94 vs.\ 80.31 on ImageNeedle; Tab.~\ref{tab:abl}(a)) while incurring less latency than a fully block-wise setup, so we used this strategy by default in experiments. We also evaluate dynamic layer-wise budgeting~\cite{flowmm, meda}, a recently proposed adaptive alternative to static allocation. In our setting, however, it underperforms static budgeting during prefill, causing a 5.63-point drop at budget 1024 (Tab.~\ref{tab:abl}(b)). This suggests that dynamic budgeting is not yet reliable in prefill, likely because attention statistics are still incomplete at that stage.


\paragraph{Compression vs.\ input reduction.}
To distinguish prefill-stage KV-cache compression from simply using fewer vision tokens, we compare our method with reducing the input resolution under the same KV-cache budget. Lowering the input itself leads to severe performance degradation (Tab.~\ref{tab:abl}(c)), whereas our method preserves high-resolution visual information while controlling memory usage through compression during prefill.

\paragraph{Block size and structural alignment.}
As shown in Tab.~\ref{tab:qwen_block}, block size has a strong impact on compression effectiveness. For Qwen2.5-VL-7B, accuracy peaks at block size 784 under a budget of 2048, which exactly matches the model’s native $28\times28$ visual tokenization. In contrast, block sizes that are misaligned with this tokenization (e.g., 512) lead to reduced robustness, explaining the larger performance drop observed for Qwen2.5-VL-7B in Tab.~\ref{tab:main_result_trimmed}. This is not merely a model-specific artifact, but also a counterexample supporting our broader argument that structural alignment matters: compression granularity must respect the spatial structure of visual representations. Taken together, these results highlight that vision-aware block design is critical for maintaining performance.

\begin{table}[t]
\centering
\resizebox{0.48\textwidth}{!}{
\setlength{\tabcolsep}{2pt}
\begin{tabular}{c|ccc}
\toprule[1.25pt]
\toprule 
Block Size (KV Budget) & ImageNeedle & Global Peak (GB) & Avg. Peak (GB) \\
\midrule
\multicolumn{4}{l}{\textbf{\textit{\small{Qwen2.5-VL-7B}}}} \\
\midrule
256 (2048) & 72.19 & 17.80 & 17.12  \\
512 (2048) & 75.31 & 18.00 & 17.19  \\
\rowcolor{gray!15}784 (2048) & 80.63 & 18.21 & 17.26  \\
1024 (2048) & 79.38 & 18.38 & 17.37  \\
\bottomrule[1.25pt] 
\end{tabular}
}
\caption{\textbf{Effect of block size} under a fixed KV-cache budget (Qwen2.5-VL-7B). Performance peaks at block size 784, which matches the model’s native visual tokenization.}
\vspace{-17pt}
\label{tab:qwen_block}
\end{table}

\section{Conclusion}
\vspace{-1pt}
We propose a prefill-aware, block-wise KV-cache compression method that significantly reduces memory use during multimodal inference. 
By compressing online during prefill, our approach maintains a nearly constant peak memory footprint under fixed KV-cache budgets while avoiding out-of-memory failures.
Extensive experiments across image and video benchmarks show that our method achieves up to $\sim$90\% cache reduction with minimal performance degradation. 

Our results highlight that memory efficiency in MLLMs is strongly influenced not only by the final cache size, but also by how visual context is processed during prefill. 
We hope this work provides a step toward scalable and memory-efficient multimodal inference under practical system constraints.

\section*{Limitations}
Our method enforces a fixed KV-cache budget during prefill and therefore introduces several natural trade-offs. Block-wise prefill processes inputs sequentially, which can increase inference latency compared to bulk execution, reflecting an inherent memory–latency trade-off; in practice, this overhead can be mitigated with hybrid execution strategies. In addition, compression effectiveness depends on alignment between block boundaries and the structure of visual representations, and query-agnostic eviction prioritizes general representational diversity rather than task-specific relevance. Finally, our approach focuses on inference-time optimization without modifying training, and models explicitly trained with prefill-stage compression may further improve robustness.

\section*{Ethical Considerations}
This work focuses on improving inference-time memory efficiency for multimodal large language models through KV-cache management. The proposed method does not introduce new model capabilities, training data, or deployment scenarios, and does not alter model behavior beyond resource usage. As such, it does not raise additional ethical concerns beyond those already associated with large language models and multimodal systems in general.

\FloatBarrier
\bibliography{custom}

\clearpage
\appendix
\appendix

\section*{\Large Appendix}

\section{Related Works}
The rapid growth of the key--value (KV) cache in long-context inference has motivated extensive research on cache compression methods~\cite{li2024snapkv, h2o, keydiff, streamingllm}, commonly categorized into quantization-, eviction-, and merging-based approaches. This work focuses on eviction-based methods, with emphasis on their limitations in multimodal large language models (MLLMs).

\paragraph{KV-Cache Eviction in LLMs.}
KV-cache eviction strategies maintain a fixed memory budget by selectively discarding tokens that are unlikely to contribute to future generation. Early methods such as StreamingLLM~\cite{streamingllm} identify \emph{attention sinks} and retain them together with a sliding window of recent tokens. More advanced approaches, including H$_\text{2}$O~\cite{h2o} and SnapKV~\cite{li2024snapkv}, leverage accumulated attention statistics to preserve \emph{heavy hitter} tokens that are frequently attended to during decoding.

Complementary query-agnostic strategies avoid reliance on a specific query signal. KeyDiff~\cite{keydiff} observes that highly attended tokens tend to be representationally diverse, and therefore retains keys that are distant from the centroid of the key distribution. Unlike query-dependent methods, such approaches enable the compressed KV cache to be reused across different queries.

\paragraph{KV-Cache Eviction in MLLMs.}
In multimodal settings, KV-cache eviction must additionally address the substantial redundancy introduced by large numbers of visual tokens. LOOK-M~\cite{lookm} exploits the tendency of MLLMs to prioritize textual tokens, selectively pruning visual tokens while preserving the text prompt. MEDA~\cite{meda} introduces layer-wise adaptive budget allocation guided by cross-modal attention entropy, allowing visually sensitive layers to retain denser representations. FlowMM~\cite{flowmm} further extends this direction by dynamically merging tokens based on cross-modal attention patterns. Together, these methods move beyond coarse window-based pruning toward more modality-aware KV-cache management, but primarily operate after the full multimodal context has been processed.

\section{General Multimodal Performance}

As discussed in Sec.~\ref{subsec:exp_setup}, we primarily focus on benchmarks that are sensitive to the scale and structure of visual tokens. To further assess the general applicability of our method, we additionally evaluate InternVL3.5-8B~\cite{wang2025internvl35} on MMMU~\cite{yue2024mmmu} and POPE~\cite{li2023pope}. As shown in Tab.~\ref{tab:general}, the small gap from the full-cache baseline indicates that our method remains robust even on more general multimodal tasks beyond benchmarks specifically designed to stress high-resolution inputs.


\section{Discussion}
The results demonstrate that controlling peak memory during the prefill stage is both feasible and critical for scaling multimodal inference to high-resolution and long-context visual inputs. 
By shifting KV-cache compression from a post-prefill operation to an online, structure-aligned process, our framework enables models to better retain visual information while operating under strict memory budgets. 
The consistent performance observed across image and video benchmarks, together with stable peak memory usage and the avoidance of out-of-memory failures, suggests that prefill-aware compression addresses a fundamentally different bottleneck than existing decoding-time methods. 
More broadly, these findings indicate that memory efficiency in MLLMs is not solely a function of final cache size, but is strongly shaped by how and when visual context is processed during inference.

\begin{table}[t]
\centering
\resizebox{0.4\textwidth}{!}{
\setlength{\tabcolsep}{2pt}
\begin{tabular}{c|cccc}
\toprule[1.25pt]
\toprule 
Method (KV Budget) & MMMU & POPE & Average & $\Delta$ \\
\midrule
\multicolumn{5}{l}{\textbf{\textit{\small{InternVL3.5-8B}}}} \\
\midrule
Full Cache   &  56.28 & 87.40 & 71.84 & - \\
\midrule
SnapKV (1024)  & 56.48 & 87.02 & 71.75 & 0.09  \\
KeyDiff (1024) & 54.20 & 88.63 & 71.42 & 0.42 \\
\bottomrule[1.25pt] 
\end{tabular}
}
\caption{\textbf{General Multimodal Performance.} Results on MMMU and POPE with InternVL3.5-8B. For fair comparison with Tab.~\ref{tab:main_result_trimmed}, we use the same increased vision token configuration as in the main experiments.}
\vspace{-5pt}
\label{tab:general}
\end{table}

\section{Use of Large Language Models}
In accordance with the ACL 2026 submission policy, we disclose that Large Language Models were used to assist in grammar correction and polishing of the writing in this paper.

\label{sec:appendix}

\end{document}